\documentclass{article}

\usepackage{arxiv}

\usepackage[utf8]{inputenc}
\usepackage[T1]{fontenc}
\usepackage{hyperref}
\usepackage{url}
\usepackage{booktabs}
\usepackage{amsfonts}
\usepackage{amsmath,amssymb}
\usepackage{algorithmic}
\usepackage{algorithm}
\usepackage{array}
\usepackage{microtype}
\usepackage{graphicx}
\usepackage[numbers,sort&compress]{natbib}
\usepackage{doi}
\usepackage{caption}
\usepackage{subcaption}
\usepackage{enumitem}
\usepackage{textcomp}
\usepackage{placeins}
\usepackage{xspace}
\usepackage[dvipsnames]{xcolor}
\usepackage{colortbl}
\usepackage{pifont}

\definecolor{ForestGreen}{RGB}{34,139,34}

\newcommand{\cmark}{\ding{51}}
\newcommand{\xmark}{\ding{55}}

\makeatletter
\DeclareRobustCommand\onedot{\futurelet\@let@token\@onedot}
\def\@onedot{\ifx\@let@token.\else.\null\fi\xspace}

\makeatother

\definecolor{gain}{HTML}{34a853}

\definecolor{lost}{HTML}{ea4335}

\renewcommand{\headeright}{arXiv Preprint}
\renewcommand{\undertitle}{arXiv Preprint}
\renewcommand{\shorttitle}{PhyUnfold-Net}

\title{PhyUnfold-Net: Advancing Remote Sensing Change Detection with Physics-Guided Deep Unfolding}
\date{}

\author{
  Zelin Lei\thanks{Equal contribution.} \\
  Xi'an Jiaotong University \\
  \texttt{thedieisnotcast@stu.xjtu.edu.cn}
  \And
  Yaoxing Ren\footnotemark[1] \\
  Xi'an Jiaotong University \\
  \texttt{renyaoxing2000@163.com}
  \And
  Jiaming Chang\thanks{Corresponding author.} \\
  Anhui University \\
  \texttt{jmchang0517@163.com}
}

\hypersetup{
  pdftitle={PhyUnfold-Net: Advancing Remote Sensing Change Detection with Physics-Guided Deep Unfolding},
  pdfsubject={Computer Science - Computer Vision and Pattern Recognition (cs.CV)},
  pdfauthor={Zelin Lei, Yaoxing Ren, Jiaming Chang},
  pdfkeywords={remote sensing, bi-temporal change detection, deep unfolding, singular-value entropy},
}

\begin{document}
\maketitle

\begin{abstract}
Bi-temporal change detection is highly sensitive to acquisition discrepancies, including illumination, seasonal, and atmospheric variations, which often cause false alarms.
We observe that genuine changes tend to exhibit higher patch-wise singular-value entropy (SVE) than unchanged or nuisance-dominated regions in the feature-difference space.
Motivated by this physical prior, we propose PhyUnfold-Net, a physics-guided deep unfolding framework that formulates change detection as an explicit decomposition problem.
The proposed Iterative Change Decomposition Module (ICDM) unrolls a multi-step solver that progressively separates mixed discrepancy features into change and nuisance components.
To stabilize this process, we introduce a Staged Separation Exploration-and-Constraint loss (S-SEC), which promotes component separation in early steps and constrains nuisance magnitude in later steps to reduce degenerate solutions.
We further design a Wavelet Spectral Suppression Module (WSSM) to suppress acquisition-induced spectral mismatch before decomposition.
Experiments on four benchmarks demonstrate consistent improvements, with PhyUnfold-Net achieving the best F1 and IoU on all four datasets among the methods compared.

\end{abstract}

\section{Introduction}
\label{sec:intro}
Change detection in bi-temporal remote sensing imagery underpins large-scale Earth observation tasks such as urban expansion monitoring, land-use analysis, and disaster assessment in dynamic environments.
Deep learning has driven steady progress, from Siamese CNNs~\citep{daudt2018fully,fang2021snunet} to Transformer-based architectures~\citep{vaswani2017attention,dosovitskiy2021image,liu2021swin} that capture long-range context.
Nevertheless, robust change detection under complex acquisition conditions remains challenging.

Images captured at different times often differ in illumination, season, and atmospheric conditions.
These acquisition discrepancies can introduce pseudo changes and obscure genuine semantic changes, making direct prediction from raw bi-temporal features unreliable.

However, most existing change detection methods suppress nuisance implicitly, which can leave residual acquisition-induced interference.
\begin{figure*}[t]
    \centering
    \includegraphics[width=\textwidth]{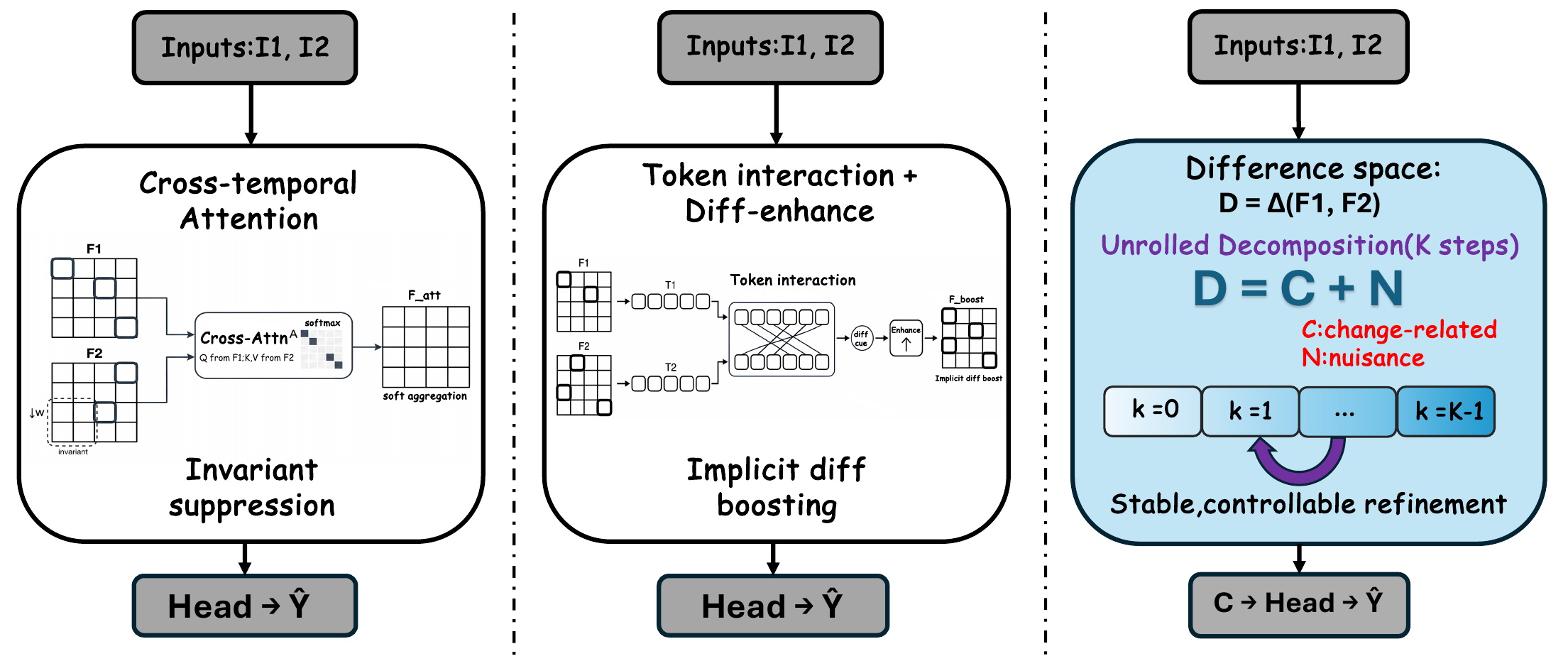}
       \caption{Comparison of change-detection paradigms. \textbf{(a)} Cross-temporal attention for invariant feature learning. \textbf{(b)} Token interaction with implicit difference enhancement. \textbf{(c)} Explicit difference-space decomposition $D=C+N$ with unrolled refinement and separation control.}

    \label{fig:paradigm}
    \vspace{-0.2cm}
\end{figure*}
Existing methods primarily employ cross-temporal attention or token interaction with difference enhancement. As shown in Fig.~\ref{fig:paradigm}(a), cross-temporal attention learns invariant correspondences between $I_1$ and $I_2$ to attenuate appearance-induced disturbances before predicting $\hat{Y}$.
Fig.~\ref{fig:paradigm}(b) illustrates token interaction with difference enhancement, which boosts discriminative discrepancies but still entangles true change with nuisance in learned features.
In contrast, as shown in Fig.~\ref{fig:paradigm}(c), our difference-space formulation first corrects the encoder features to obtain $(\hat{F}_1,\hat{F}_2)$ and then decomposes $D=\Delta(\hat{F}_1,\hat{F}_2)$ into a change component $C$ and a nuisance component $N$ through an unrolled $K$-step solver. The final prediction is produced from $C$.

In this paper, we present \textbf{PhyUnfold-Net}, a physics-guided deep unfolding framework that casts bi-temporal change detection as a structured decomposition of the feature-difference map $D$ into a change component $C$ and a nuisance component $N$.
Our method builds on an unrolled $K$-step solver~\citep{gregor2010learning,monga2021algorithm} in which residual-driven updates progressively refine $(C,N)$.
SVE-guided gating, staged regularization, and wavelet-based spectral suppression jointly strengthen the decomposition and reduce acquisition-induced nuisance, yielding a robust and interpretable change-detection model.
Our main contributions are summarized as follows:
\begin{enumerate}[leftmargin=*,itemsep=2pt]
    \item We introduce \textbf{PhyUnfold-Net}, a physics-guided deep unfolding framework that formulates bi-temporal change detection as an explicit difference-space decomposition of $D$ into $(C,N)$ with clear physical interpretability.
    \item We design a memory-enhanced unrolled solver, \textbf{ICDM}, with SVE-gated residual injection to stabilize refinement in nuisance-dominated regions while preserving fine structures.
    \item We develop \textbf{S-SEC}, a staged regularization scheme that couples margin-based component separation with bounded nuisance-energy constraints to reduce degenerate decompositions.
    \item We incorporate \textbf{WSSM} to align multi-frequency structures across time via wavelet-based spectral suppression, thereby suppressing acquisition-induced variations and improving robustness.
\end{enumerate}

\section{Related Work}
\label{sec:related}

\subsection{Remote Sensing Change Detection}
Recent studies have advanced remote sensing change detection from several complementary perspectives. Convolutional architectures typically use hierarchical feature extraction to compare bi-temporal images, but their performance can deteriorate when changes occur at diverse spatial scales or when low-level appearance variations obscure high-level land-cover transitions. Multi-scale feature pyramids address this limitation by aggregating representations across resolution levels~\citep{lin2017feature}. Relation-aware aggregation modules further improve contextual modeling by strengthening interactions among spatial regions~\citep{chen2023saras}. Attention mechanisms allow networks to emphasize informative locations, temporal correspondences, and semantically meaningful feature channels~\citep{chen2020stanet,han2023hanet}. Lightweight models reduce computational and memory costs, facilitating deployment in resource-constrained settings while retaining competitive accuracy~\citep{codegoni2023tinycd}.

Modern sequence-modeling architectures have further reshaped the field~\citep{vaswani2017attention,dosovitskiy2021image,liu2021swin,xie2021segformer}. Transformer-based change-detection methods model bi-temporal interactions and semantic correspondences across images~\citep{chen2021remote}, while state-space models capture long-range dependencies with favorable computational efficiency~\citep{zhao2024rsmamba,ding2024joint}.

Despite these advances, most methods still model nuisance variations only implicitly. PhyUnfold-Net instead uses singular-value entropy to characterize the structural complexity of changed and nuisance-dominated regions, supporting explicit change--nuisance separation within a lightweight architecture.

\subsection{Deep Unfolding Networks}
Deep unfolding maps iterative solvers into trainable networks with step-wise refinement~\citep{beck2009fista}.
By embedding algorithm-inspired update rules, unrolled architectures provide a principled way to inject priors, control the refinement trajectory, and share parameters across steps.
Unrolling has shown clear benefits in inverse problems~\citep{gregor2010learning,zhang2018ista,monga2021algorithm}.
However, residual updates may amplify errors when intermediate states are uncertain, and repeated transformations can degrade fine structural details.
Compact recurrent memory offers a principled remedy by preserving useful information across steps under long-horizon refinement~\citep{cho2014learning,shi2015convolutional}. 

In this work, we introduce a deep unfolding architecture that decomposes discrepancy features into change and nuisance components. We further introduce a Shared-Parameter Memory Unit (SPMU) to stabilize the unfolding process.

\section{Preliminaries}
\label{sec:prelim}

\noindent\textbf{Notation and Problem Setting.}
Given a bi-temporal image pair $(I_1, I_2)$ with $I_t \in \mathbb{R}^{H \times W \times c}$ ($t\in\{1,2\}$), the goal is to predict a binary change mask $Y \in \{0,1\}^{H \times W}$.
A Siamese feature extractor $E(\cdot)$ maps each image to a feature map:
\begin{equation}
F_t = E(I_t) \in \mathbb{R}^{H'\times W'\times d}, \qquad t\in\{1,2\}.
\label{eq:pre_feat}
\end{equation}
Let $\hat{F}_t$ denote the corresponding feature after spectral correction; the identity case $\hat{F}_t=F_t$ recovers a formulation without correction. We then construct the mixed discrepancy feature
\begin{equation}
D = \Delta(\hat{F}_1,\hat{F}_2) \in \mathbb{R}^{H'\times W'\times d},
\label{eq:pre_diff}
\end{equation}
where $\Delta$ denotes the feature-difference operator. Instead of predicting directly from $D$, we reformulate the problem as
\begin{equation}
D = C + N,
\label{eq:pre_decomp}
\end{equation}
where $C$ and $N$ denote the change-related and nuisance components, respectively.

To regularize the decomposition, we formulate the objective as
\begin{equation}
\min_{C,N}\;\; 
\underbrace{\mathcal{L}_{\mathrm{rec}}(C,N;D)}_{\text{data fidelity}}
+\lambda_C \mathcal{R}_C(C)
+\lambda_N \mathcal{R}_N(N),
\label{eq:pre_var}
\end{equation}
where $\mathcal{L}_{\mathrm{rec}}(C,N;D)=\|D-(C+N)\|_1$, and $\mathcal{R}_C$ and $\mathcal{R}_N$ are generic priors weighted by $\lambda_C$ and $\lambda_N$, respectively.
ICDM unrolls a solver for \eqref{eq:pre_var} into $K$ learnable steps to iteratively refine $(C,N)$.

\noindent\textbf{Unrolled Refinement Operator.}
Let $(C^k,N^k)$ denote the refinement state at step $k$.
The learned refinement operator is
\begin{equation}
(C^{k+1},N^{k+1})=\mathcal{T}_{\theta_k}(C^k,N^k;D), \qquad k=0,\ldots,K-1,
\label{eq:pre_T}
\end{equation}
where $\mathcal{T}_{\theta_k}$ is parameterized by neural modules, and parameters can be shared across steps.
This formulation turns iterative refinement into a learnable operator with step-wise control.

\noindent\textbf{Patch-wise singular-value entropy.}
\label{sec:prelim_sve}
We quantify local spectral complexity via patch-wise singular-value entropy (SVE).
Given a feature map $X \in \mathbb{R}^{d\times H'\times W'}$, we partition its spatial domain into $n_p$ non-overlapping square patches of side length and stride $p$:
\[
\mathcal{P}_j=\{(u,v)\mid u\in[u_j,u_j+p-1],\ v\in[v_j,v_j+p-1]\},
\]
and denote the corresponding cropped tensor by $X_{\mathcal{P}_j}\in\mathbb{R}^{d\times p\times p}$.
We reshape each patch into a channel--spatial matrix
\begin{equation}
M_j=\mathrm{reshape}(X_{\mathcal{P}_j})\in\mathbb{R}^{d\times L},\qquad L=p^2.
\label{eq:pre_M}
\end{equation}
Let the singular values of $M_j$ be $\boldsymbol{\sigma}_j=[\sigma_{j,1},\ldots,\sigma_{j,r_j}]$ with $r_j=\min(d,L)$.
Let $a_j=\sum_{\ell=1}^{r_j}\sigma_{j,\ell}$. To make the definition valid for an all-zero patch, we define the normalized spectrum by
\begin{equation}
q_{j,i}=
\begin{cases}
\sigma_{j,i}/a_j, & a_j>0,\\
0, & a_j=0,
\end{cases}
\qquad i=1,\ldots,r_j,
\label{eq:pre_p_patch}
\end{equation}
and compute the patch-wise SVE as
\begin{equation}
s_j=-\sum_{i=1}^{r_j} q_{j,i}\log q_{j,i},
\label{eq:pre_sve_patch}
\end{equation}
with the convention $0\log 0=0$. Thus, an all-zero patch has zero SVE without requiring division by zero. Finally, we form a dense map $S\in\mathbb{R}^{H'\times W'}$ by assigning $s_j$ to all pixels in $\mathcal{P}_j$.

\section{The Proposed Method}
\label{sec:method}

\begin{figure*}[t]
    \centering
    \includegraphics[width=\textwidth]{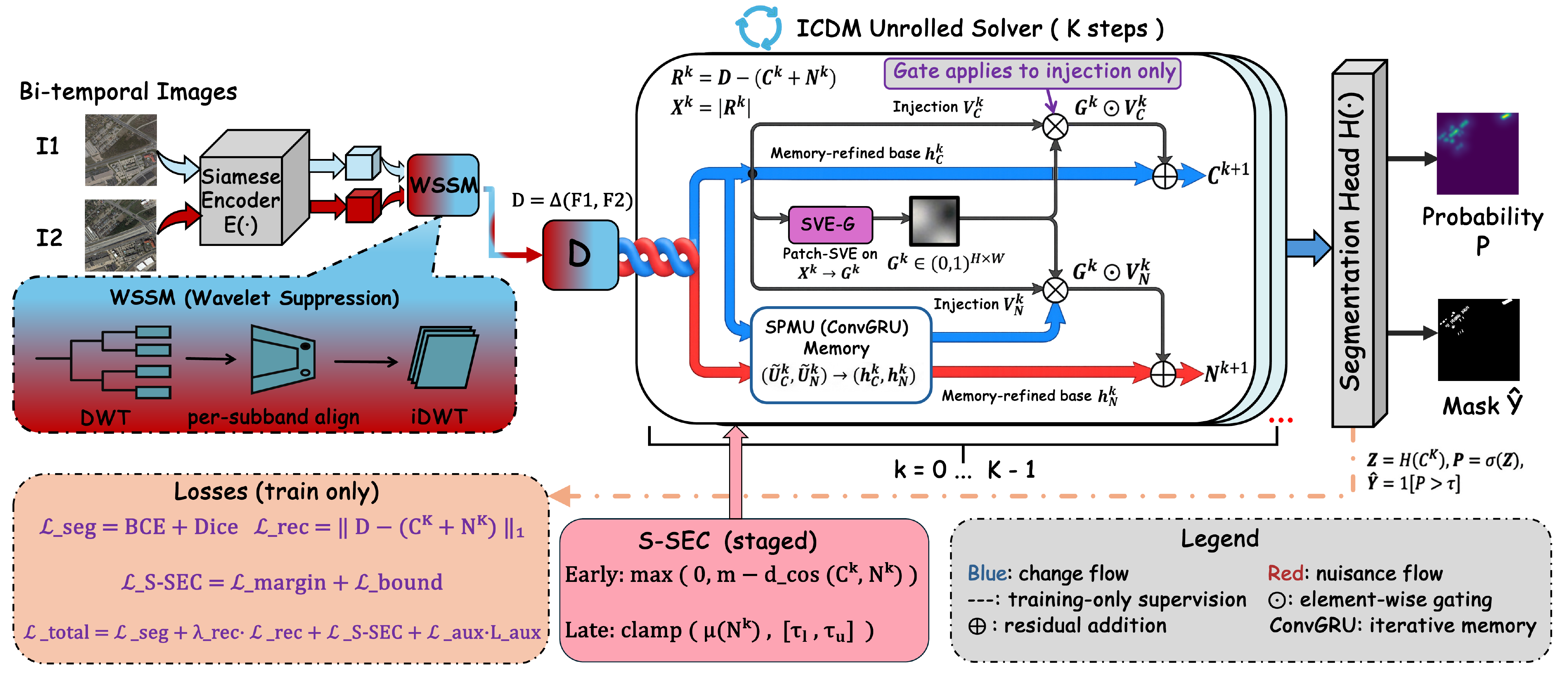}
         \caption{Overview of PhyUnfold-Net. WSSM suppresses acquisition-induced variations and produces corrected features $(\hat{F}_1,\hat{F}_2)$ before the difference map $D=\Delta(\hat{F}_1,\hat{F}_2)$ is formed. ICDM then unrolls a $K$-step solver to refine the decomposition $D=C+N$, and the segmentation head predicts $\hat{Y}$ from $C^{K}$. Training-only losses are shown in the lower panels.}
   
    \label{fig:overview}
       \vspace{-0.2cm}
\end{figure*}

We illustrate the overall pipeline of PhyUnfold-Net in Fig.~\ref{fig:overview}.
Given bi-temporal images $(I_1,I_2)$, a Siamese encoder $E(\cdot)$ extracts feature fields $(F_1,F_2)$.
To mitigate acquisition-induced appearance discrepancies, we first apply WSSM as a wavelet-domain suppression module at stages 2--4. We then fuse the corrected stage-3 and stage-4 features to form the difference map $D=\Delta(\hat{F}_1,\hat{F}_2)$.
PhyUnfold-Net models $D$ as an explicit decomposition $D=C+N$ and performs $K$-step unrolled refinement via ICDM; residual-driven updates progressively refine $(C,N)$ under spectral-aware gating and lightweight recurrent memory.
The final prediction is produced from the change component $C^{K}$ by a segmentation head, and training further incorporates reconstruction and staged regularization terms as indicated in the figure.

\subsection{Memory-Enhanced \textbf{ICDM} with SVE-Gated Residual Injection}
\label{sec:method_icdm}
\label{sec:method_sveg}

Given a bi-temporal pair $(I_1,I_2)$, a Siamese encoder $E(\cdot)$ extracts features $F_t=E(I_t)$ for $t\in\{1,2\}$. WSSM produces corrected features $\hat{F}_t$, from which we construct the discrepancy field $D=\Delta(\hat{F}_1,\hat{F}_2)$. In our implementation, $\Delta(A,B)=B-A$ after multi-scale fusion.
Because $D$ contains both genuine change cues and acquisition-induced perturbations, we decompose it explicitly as
\begin{equation}
D = C + N,
\label{eq:method_decomp}
\end{equation}
in which $C$ accumulates change-related content and $N$ absorbs nuisance variations in a complementary manner across steps.

This design is motivated by the observation that acquisition-induced nuisance often appears as locally uncertain mismatch patterns. ICDM therefore performs a $K$-step unfolding process.
We initialize $C^{0}=\mathbf{0}$ and $N^{0}=D$ and progressively extract change evidence from the initially nuisance-dominated representation.
At iteration $k$, we define the decomposition residual as
\begin{equation}
R_{\mathrm{res}}^{k} = D-(C^{k}+N^{k}),
\label{eq:icdm_residual}
\end{equation}
which quantifies the mismatch between the observed difference map and the current reconstruction.
Rather than updating the two branches independently, ICDM predicts coupled directions from their joint context:
\begin{align}
\Delta C^{k} = \Phi_C\!\big([C^{k},N^{k},R_{\mathrm{res}}^{k}]\big),
\\
\Delta N^{k} = \Phi_N\!\big([C^{k},N^{k},R_{\mathrm{res}}^{k}]\big).
\label{eq:icdm_delta}
\end{align}
Here, $[\cdot]$ denotes channel concatenation, and $\Phi_C,\Phi_N$ are lightweight learnable operators.
The update consists of a base correction along the predicted directions and a residual-reinjection path for unresolved discrepancies:
\begin{equation}
\tilde U_C^{k} = \alpha_k\,\Delta C^{k},\qquad
\tilde U_N^{k} = \beta_k\,\Delta N^{k},
\label{eq:icdm_base_update}
\end{equation}
\begin{equation}
V_C^{k} = \gamma_k\,\Psi_C(R_{\mathrm{res}}^{k}),\qquad
V_N^{k} = \gamma_k\,\Psi_N(R_{\mathrm{res}}^{k}).
\label{eq:icdm_res_inject}
\end{equation}
The learnable scalars $\alpha_k,\beta_k,\gamma_k$ control step amplitudes, and $\Psi_C,\Psi_N$ are $1{\times}1$ projections.
To prevent long-horizon drift while preserving fine structures, we first form provisional states using only the base increments:
\begin{equation}
\tilde C^{k+1} = C^{k}+\tilde U_C^{k},\qquad
\tilde N^{k+1} = N^{k}+\tilde U_N^{k}.
\label{eq:icdm_provisional}
\end{equation}
We then propagate compact recurrent memories:
\begin{align}
h_C^{k} = \mathrm{Mem}_C\!\big(\tilde C^{k+1},h_C^{k-1}\big),
\\
h_N^{k} = \mathrm{Mem}_N\!\big(\tilde N^{k+1},h_N^{k-1}\big),
\label{eq:icdm_memory}
\end{align}
with initialization $h_C^{-1}=h_N^{-1}=0$.
Each branch uses a \textbf{Shared-Parameter Memory Unit (SPMU)} with lightweight gated convolutional state transitions inspired by GRU and ConvLSTM~\citep{cho2014learning,shi2015convolutional}. The parameters of each SPMU are shared across unrolling steps, providing recurrent context without increasing the parameter count with $K$.

The same physical prior further suggests that nuisance-dominated mismatch is locally uncertain.
Accordingly, residual reinjection should be spatially moderated rather than uniformly applied.
We derive a local uncertainty signal from residual magnitude at each spatial location,
\begin{equation}
X^{k} = \left|R_{\mathrm{res}}^{k}\right|\in\mathbb{R}^{d\times H'\times W'},
\label{eq:method_gate_input}
\end{equation}
and reduce channels before entropy estimation:
\begin{equation}
\bar X^k = \Pi(X^k)\in\mathbb{R}^{d_{\text{sve}}\times H'\times W'},\qquad d_{\text{sve}}\ll d.
\label{eq:method_channel_reduce}
\end{equation}
Channel reduction is implemented by a learnable $1{\times}1$ projection $\Pi(\cdot)$.
Patch-wise SVE on $\bar X^k$ (Sec.~\ref{sec:prelim_sve}) produces an entropy map $S^k\in\mathbb{R}^{H'\times W'}$, which a learnable mapping converts into an adaptive residual-reinjection gate:
\begin{equation}
G^{k}=\operatorname{sigmoid}\!\big(\phi(S^{k})\big),
\label{eq:method_gate_map}
\end{equation}
where $\phi(\cdot)$ transforms $S^k$ into $G^k\in(0,1)^{H'\times W'}$.
The final update applies this gate to the residual-injection paths:
\begin{align}
C^{k+1}=h_C^{k}+\big(G^{k}\otimes \mathbf{1}\big)\odot V_C^{k},
\\
N^{k+1}=h_N^{k}+\big(G^{k}\otimes \mathbf{1}\big)\odot V_N^{k}.
\label{eq:icdm_final_update}
\end{align}
In Eq.~\eqref{eq:icdm_final_update}, $\odot$ denotes element-wise multiplication, and $G^k\otimes\mathbf{1}$ broadcasts the 2D gate over channels.
This gate modulates residual corrections spatially, preventing uniform reinjection in heterogeneous regions.
Repeating this refinement for $K$ iterations yields $(C^K,N^K)$ as the final decomposed representation.

Given the final change-related component $C^{K}$, a lightweight segmentation head $H(\cdot)$ predicts logits and probabilities as:
\begin{equation}
Z = H(C^{K}), \qquad P = \sigma(Z), \qquad \hat{Y} = \mathbb{I}[P > \tau],
\label{eq:method_pred_head}
\end{equation}
with sigmoid activation $\sigma(\cdot)$ and decision threshold $\tau$. During training, segmentation supervision is applied to the bilinear upsampled $P$, while $\hat{Y}$ is used only for inference.
\subsubsection{Assumption}

Assume that there exists a contraction factor $\rho\in(0,1)$ such that, for every $k\ge 1$, the residuals satisfy
\begin{equation}
\|R_{\mathrm{res}}^{k+1}\|_F \;\leq\; \rho\,\|R_{\mathrm{res}}^{k}\|_F + \delta_k,
\label{eq:supp_contract}
\end{equation}
where $\delta_k\geq 0$ is a perturbation term associated with SVE gating and memory updates, and $\sum_{k=1}^{\infty}\delta_k < \infty$.
Note that Eq.~\eqref{eq:supp_contract} does not require monotonic decrease at every step. The initialization transition $k=0\!\to\!1$ is treated as a transient, which is not constrained by Eq.~\eqref{eq:supp_contract}.

\subsubsection{Justification}
The assumption is motivated by three architectural choices: (i)~the base update $(\tilde{U}_C^k,\tilde{U}_N^k)$ depends explicitly on $R_{\mathrm{res}}^k$; (ii)~the sigmoid gate bounds the multiplicative gating coefficient between zero and one; and (iii)~the recurrent memory propagates information across steps and reduces abrupt state changes. These choices motivate, but do not by themselves guarantee, contraction because the learned projections and memory operators remain unconstrained. Accordingly, Eq.~\eqref{eq:supp_contract} is a sufficient condition for the following residual bound, not an unconditional convergence guarantee for every trained model.

\subsubsection{Convergence Analysis}
\label{prop:convergence}
Under the above assumption, the decomposition residual satisfies
\begin{equation}
\|R_{\mathrm{res}}^{K}\|_F \;\leq\; \rho^{K-1}\|R_{\mathrm{res}}^{1}\|_F + \sum_{j=1}^{K-1}\rho^{K-1-j}\,\delta_j,\qquad K\ge 2.
\label{eq:supp_bound}
\end{equation}
Here, $R_{\mathrm{res}}^{k}=D-(C^{k}+N^{k})$ denotes the decomposition residual at step $k$, $K\in\mathbb{N}$ is the total number of iterations, $\rho\in(0,1)$ is the contraction factor, $\delta_j\ge 0$ is the perturbation term (with $\sum_{j=1}^{\infty}\delta_j<\infty$), and $\|\cdot\|_F$ denotes the Frobenius norm.
(For $K{=}1$ the bound holds trivially as an identity.)
If the recursion were extended indefinitely under the same contraction factor and summable perturbations, then
\begin{equation}
\lim_{K\to\infty}\|R_{\mathrm{res}}^{K}\|_F = 0.
\label{eq:supp_limit}
\end{equation}
This limit is a theoretical extrapolation of the finite-step recursion rather than a deployment-time guarantee.
Our model uses fixed $K{=}3$ with step-specific parameters $\{\theta_k\}_{k=0}^{2}$, apart from the memory-unit parameters shared across steps; the operative bound is Eq.~\eqref{eq:supp_bound} evaluated at $K{=}3$.

\subsubsection{Proof}
Applying inequality~\eqref{eq:supp_contract} recursively:
\begin{align}
\|R_{\mathrm{res}}^{2}\|_F &\leq \rho\|R_{\mathrm{res}}^{1}\|_F + \delta_1, \notag\\
\|R_{\mathrm{res}}^{3}\|_F &\leq \rho^2\|R_{\mathrm{res}}^{1}\|_F + \rho\delta_1 + \delta_2, \notag\\
&\;\;\vdots \notag\\
\|R_{\mathrm{res}}^{K}\|_F &\leq \rho^{K-1}\|R_{\mathrm{res}}^{1}\|_F + \sum_{j=1}^{K-1}\rho^{K-1-j}\delta_j.
\label{eq:supp_recursive}
\end{align}
This establishes~\eqref{eq:supp_bound}.
For the limit, the first term vanishes because $\rho<1$. Summability implies $\delta_k\to0$, and the convolution of the vanishing sequence $\{\delta_k\}$ with the geometric kernel $\{\rho^k\}$ also tends to zero. Hence $\|R_\mathrm{res}^K\|_F\to0$, which proves~\eqref{eq:supp_limit} and implies $(C^K+N^K)\to D$.

We equip the product space $\mathbb{R}^{d\times H'\times W'}\times\mathbb{R}^{d\times H'\times W'}$ with the standard Euclidean norm $\|(A,B)\|^2=\|A\|_F^2+\|B\|_F^2$, and define $\operatorname{dist}(z,\mathcal{M})=\inf_{z'\in\mathcal{M}}\|z-z'\|$.

\subsubsection{Corollary}
Define $z^k=(C^k,N^k)$ and
$
\mathcal{M}=\{(C,N): C+N=D\}.
$
Under the above assumption,
\begin{equation}
\operatorname{dist}(z^K,\mathcal{M})=\frac{1}{\sqrt{2}}\|R_{\mathrm{res}}^K\|_F.
\label{eq:supp_dist_identity}
\end{equation}
Consequently,
\begin{equation}
\lim_{K\to\infty}\operatorname{dist}(z^K,\mathcal{M})=0.
\label{eq:supp_dist_limit}
\end{equation}
This result establishes convergence to the feasible decomposition set, but it does not imply convergence to a unique pair $(C,N)$.

\subsection{Training Objective and Inference}
\label{sec:method_obj}

Explicit decomposition may admit trivial solutions without additional constraints.
We therefore adopt a \textbf{Staged Separation Exploration-and-Constraint objective (S-SEC)} that promotes early separation between components and bounds nuisance energy in later steps.
The separation measure is defined as
\begin{equation}
d_{\mathrm{sep}}(C^k,N^k) = 1-\frac{\langle \mathrm{vec}(C^k),\,\mathrm{vec}(N^k)\rangle}{\|\mathrm{vec}(C^k)\|_2\,\|\mathrm{vec}(N^k)\|_2+\epsilon}.
\label{eq:method_d}
\end{equation}
This metric operates on vectorized features via $\mathrm{vec}(\cdot)$ and the inner product $\langle\cdot,\cdot\rangle$, with $\epsilon>0$ ensuring numerical stability.
In early iterations, we enforce margin-based separation using the following loss:
\begin{equation}
\mathcal{L}_{\mathrm{exp}} 
= \sum_{k\in\mathcal{K}_\mathrm{e}} 
\max\big(0, m - d_{\mathrm{sep}}(C^k,N^k)\big),
\label{eq:method_exp}
\end{equation}
where $m$ denotes the margin, and $\mathcal{K}_\mathrm{e}$ contains the early unrolling steps used for component exploration.
In later iterations, we constrain the mean absolute magnitude of $N^k$:
\begin{equation}
\mu(N^k)=\frac{1}{H'\,W'\,d}\sum_{u=1}^{H'}\sum_{v=1}^{W'}\sum_{c=1}^{d}\left|N^k_{c}(u,v)\right|.
\label{eq:method_mu}
\end{equation}
The statistic $\mu(N^k)$ averages absolute nuisance magnitude over all $H'\times W'$ spatial locations and all $d$ channels at each step.
We further enforce a bounded range to discourage nuisance-energy collapse and uncontrolled growth:
\begin{equation}
\mathcal{L}_{\mathrm{con}}
= \sum_{k\in\mathcal{K}_\mathrm{l}}
\Big[
\max(0, \mu(N^k) - \tau_u)
+ \max(0, \tau_\ell - \mu(N^k))
\Big].
\label{eq:method_con}
\end{equation}
The later-step indices are collected in $\mathcal{K}_\mathrm{l}$, and $\tau_\ell,\tau_u$ define the nuisance-energy bounds. The sets $\mathcal{K}_\mathrm{e}$ and $\mathcal{K}_\mathrm{l}$ are disjoint subsets of $\{1,\ldots,K\}$ and exclude the initialization step $k=0$.
The full staged regularizer is defined as:
\begin{equation}
\mathcal{L}_{\mathrm{S\text{-}SEC}}
= \lambda_{\mathrm{e}}\mathcal{L}_{\mathrm{exp}}
+ \lambda_{\mathrm{c}}\mathcal{L}_{\mathrm{con}}.
\label{eq:method_eamsec}
\end{equation}
The coefficients $\lambda_{\mathrm{e}}$ and $\lambda_{\mathrm{c}}$ control the strengths of the exploration and constraint terms across unrolling stages.

The overall objective combines segmentation supervision with decomposition regularization:
\begin{equation}
\begin{aligned}
\mathcal{L}
= &\mathcal{L}_{\mathrm{seg}}\!\big(\mathrm{Up}(P), Y\big)
+ \lambda_{\mathrm{rec}}\mathcal{L}_{\mathrm{rec}}(C^K,N^K;D)
\\
+ &\mathcal{L}_{\mathrm{S\text{-}SEC}}
+ \lambda_{\mathrm{aux}}\mathcal{L}_{\mathrm{aux}}.
\end{aligned}
\label{eq:method_loss}
\end{equation}
In the final objective, $\mathcal{L}_{\mathrm{seg}}$ combines BCE and Dice losses~\citep{milletari2016vnet} computed from $(\mathrm{Up}(P),Y)$, $\mathcal{L}_{\mathrm{rec}}(C^K,N^K;D)=\|D-(C^K+N^K)\|_1$ penalizes final-step reconstruction error, and $\mathcal{L}_{\mathrm{aux}}$ denotes auxiliary segmentation supervision applied to intermediate predictions. The coefficients $\lambda_{\mathrm{rec}}$ and $\lambda_{\mathrm{aux}}$ set the corresponding weights.
At test time, we run the same $K$ ICDM steps to obtain $C^K$ and produce the final mask using Eq.~\eqref{eq:method_pred_head}.

\subsection{Wavelet Spectral Suppression Module}
\label{sec:method_wssm}

Before unrolled decomposition, we suppress acquisition-induced nuisance at the feature level.
Illumination and atmospheric variations primarily affect the low-frequency envelope of feature maps, while genuine structural changes manifest across multiple frequency bands.
Given encoder features $F_t^{(l)}$ at stage $l$ ($l \in \{2,3,4\}$), we apply a single-level 2D discrete wavelet transform (DWT) with Haar wavelets. Let $\mathcal{B}=\{\mathrm{LL},\mathrm{LH},\mathrm{HL},\mathrm{HH}\}$ denote the four subbands:
\begin{equation}
\{S_{t,b}^{(l)}\}_{b\in\mathcal{B}} = \mathrm{DWT}(F_t^{(l)}), \qquad t \in \{1,2\}.
\label{eq:wssm_dwt}
\end{equation}
Here, $t$ indexes the temporal branch, $l$ indexes the encoder stage, and $\mathrm{LL}$ denotes the low-frequency approximation; the other elements of $\mathcal B$ denote the three detail subbands.
For each $b\in\mathcal{B}$, we compute branch-wise cross-temporal residuals and suppress them by
\begin{align}
\hat{S}_{1,b}^{(l)} = S_{1,b}^{(l)} - \eta_b \Psi_b\!\big(S_{1,b}^{(l)} - S_{2,b}^{(l)}\big),
\\
\hat{S}_{2,b}^{(l)} = S_{2,b}^{(l)} + \eta_b \Psi_b\!\big(S_{1,b}^{(l)} - S_{2,b}^{(l)}\big).
\label{eq:wssm_align}
\end{align}
After alignment, $\hat{S}_{t,b}^{(l)}$ is the corrected subband feature, $\eta_b$ controls the suppression strength, and $\Psi_b$ is a learnable $1{\times}1$ projection for subband $b$.
The low-frequency subband $\mathrm{LL}$ receives stronger initial suppression, while the detail subbands use milder corrections to preserve structural edges.
The aligned subbands are reconstructed via the inverse DWT:
\begin{equation}
\hat{F}_t^{(l)} = \mathrm{IDWT}\!\big(\{\hat{S}_{t,b}^{(l)}\}_{b\in\mathcal{B}}\big).
\label{eq:wssm_idwt}
\end{equation}
This gives $\hat{F}_t^{(l)}$, the reconstructed and nuisance-suppressed feature map at stage $l$.
The cleaned features $\hat{F}_t^{(l)}$ replace $F_t^{(l)}$ before the difference operator $\Delta$.
WSSM adds limited overhead (four $1{\times}1$ convolutions per stage) and is applied at stages 2--4 of the Siamese encoder in all experiments.

\section{Experiments}
\label{sec:experiments}

\textbf{Physical Prior Validation.}
We first verify whether the proposed physical prior can emerge in a standard backbone without any explicit physics-guided module.

We train a U-Net~\citep{ronneberger2015unet} on LEVIR-CD and track the SVE of difference features $D{=}F_2{-}F_1$ at the deepest encoder stage (Encoder-4) and three decoder stages (Decoder-1, Decoder-2, and Decoder-3).
Features are partitioned into $8{\times}8$ spatial patches and grouped by ground truth.

\begin{figure}[!t]
\centering
\includegraphics[width=1.0\linewidth]{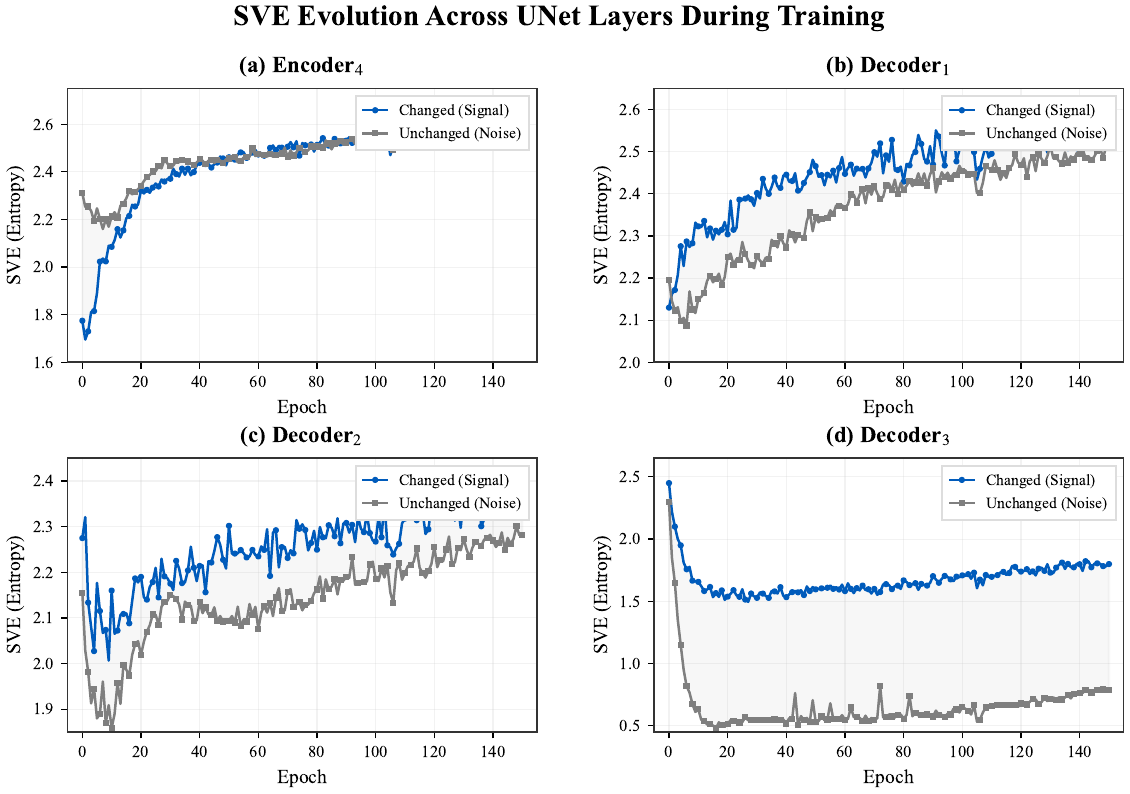}
\caption{SVE evolution across U-Net layers during training at Encoder-4, Decoder-1, Decoder-2, and Decoder-3.}
\label{fig:unet_sve}
\end{figure}
\begin{figure}
    \centering
    \includegraphics[width=1.0\linewidth]{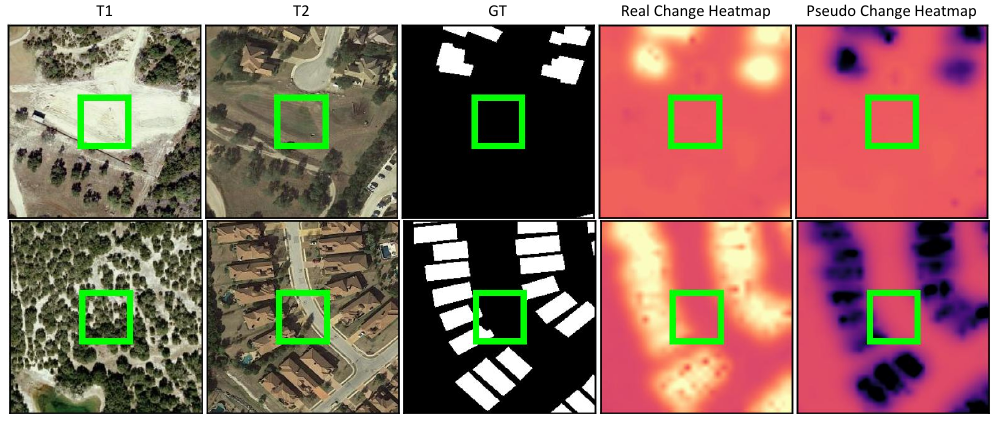}
    \caption{Qualitative analysis of the physical prior under vegetation and land-cover variations.}
    \label{fig:sub_a}
\end{figure}
\begin{figure}
    \centering
    \includegraphics[width=1.0\linewidth]{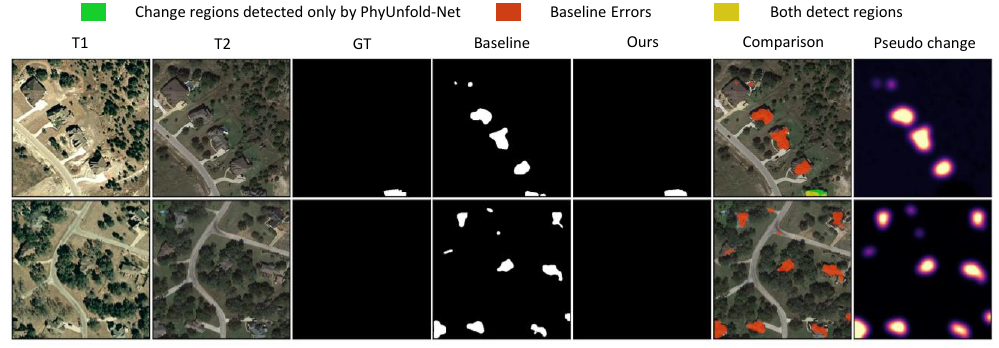}
    \caption{Robustness of the physical prior under severe illumination interference.}
    \label{fig:sub_b}
\end{figure}

Figure~\ref{fig:unet_sve} shows a clear trend: changed and unchanged patches are close in Encoder-4 and Decoder-1, then separate in Decoder-2, and diverge most in Decoder-3.
The unchanged curve keeps decreasing while the changed curve remains in a higher band, indicating that decoding suppresses nuisance complexity while preserving structurally rich change cues.
This behavior supports the physical prior used in our method under progressive decoding refinement.

We further evaluate the robustness of the physical prior. As shown in Fig.~\ref{fig:sub_a}, ICDM suppresses pseudo changes caused by vegetation growth on smooth terrain and by the conversion of forested areas into roads. Figure~\ref{fig:sub_b} shows that ICDM also remains effective when illumination variations overlap with building regions, whereas the baseline produces numerous errors.
\subsection{Datasets and Evaluation Metrics}
\label{subsec:datasets}

\noindent\textbf{LEVIR-CD}~\citep{chen2020stanet} has 637 pairs. We use the default 445 / 64 / 128 split (train / val / test), yielding 7,120 / 1,024 / 2,048 non-overlapping $256{\times}256$ patches.

\noindent\textbf{LEVIR-CD+}~\citep{shi2022deeply} has 985 pairs. We follow the official 65\% / 35\% train/test split (no official validation set).

\noindent\textbf{WHU-CD}~\citep{ji2018fully} has no official split. Following common practice~\citep{chen2021remote}, we use non-overlapping $256{\times}256$ crops with a random 6,096 / 762 / 762 split (train / val / test).

\noindent\textbf{S2Looking}~\citep{shen2021s2looking} has 5,000 pairs and 65,920 annotated change instances. We use the official 70\% / 10\% / 20\% split (3,500 / 500 / 1,000 for train / val / test).

\noindent\textbf{Metrics.} For all results produced in this work, we accumulate true positives (TP), false positives (FP), and false negatives (FN) over all test pixels and compute
\begin{equation}
\mathrm{Pre.}=\frac{\mathrm{TP}}{\mathrm{TP}+\mathrm{FP}},\qquad
\mathrm{Rec.}=\frac{\mathrm{TP}}{\mathrm{TP}+\mathrm{FN}},
\end{equation}
\begin{equation}
\mathrm{F1}=\frac{2\,\mathrm{Pre.}\,\mathrm{Rec.}}{\mathrm{Pre.}+\mathrm{Rec.}},\qquad
\mathrm{IoU}=\frac{\mathrm{F1}}{2-\mathrm{F1}}.
\end{equation}
F1 and IoU are the primary metrics~\citep{chen2021remote}. To keep the author-result rows internally consistent, their F1 and IoU values are derived from the tabulated precision and recall using the identities above and then rounded to two decimal places. Results for comparison methods are reproduced from their cited publications.

\subsection{Implementation Details}
\label{subsec:implementation}

We use ResNet-18~\citep{he2016deep} pretrained on ImageNet~\citep{deng2009imagenet} as the Siamese encoder. WSSM operates at stages 2--4, and ICDM performs $K{=}3$ unrolling steps on the difference map $D$ constructed from fused stage-3 and stage-4 features. We use $\mathcal{K}_{\mathrm e}{=}\{1\}$ and $\mathcal{K}_{\mathrm l}{=}\{2,3\}$, excluding the initialization step. PhyUnfold-Net and its ablation variants are trained for 200 epochs with AdamW~\citep{loshchilov2019decoupled,kingma2015adam} ($\beta_1{=}0.9$, $\beta_2{=}0.999$, weight decay $10^{-4}$), a batch size of 8, and a polynomial learning-rate schedule (initial rate $6{\times}10^{-4}$, power 0.9). We use automatic mixed precision, an exponential moving average with decay 0.995, random flips, and rotations within $\pm15^{\circ}$. The inference threshold is $\tau{=}0.4$. All experiments are conducted on a single NVIDIA RTX 4090 GPU.

\subsection{Comparison with State-of-the-Art Methods}
\label{subsec:main_results}

We compare PhyUnfold-Net with representative state-of-the-art methods, including FC-Siam-Di~\citep{daudt2018fully}, DTCDSCN~\citep{liu2020building}, DSIFN~\citep{zhang2020ifnet}, SNUNet~\citep{fang2021snunet}, STANet~\citep{chen2020stanet}, BIT~\citep{chen2021remote}, RS-Mamba~\citep{zhao2024rsmamba}, ChangeDA~\citep{meng2025changeda}, and SPMNet~\citep{wang2025spmnet}.

\begin{table}[!t]
\centering
\caption{Results on \textbf{LEVIR-CD}. Best in \textbf{bold}, second-best \underline{underlined}.}
\label{tab:levir}
\setlength{\tabcolsep}{5pt}
\resizebox{\columnwidth}{!}{%
\begin{tabular}{l|c|cccc}
\hline
Method & Year & Pre.(\%) & Rec.(\%) & F1(\%) & IoU(\%) \\
\hline
FC-Siam-Di~\citep{daudt2018fully} & 2018 & 89.53 & 83.31 & 86.31 & 75.92 \\
DTCDSCN~\citep{liu2020building} & 2021 & 88.54 & 86.13 & 87.32 & 77.49 \\
STANet~\citep{chen2020stanet} & 2020 & 89.06 & 85.80 & 87.40 & 77.67 \\
DSIFN~\citep{zhang2020ifnet} & 2020 & 90.15 & 87.03 & 88.56 & 79.52 \\
SNUNet~\citep{fang2021snunet} & 2022 & 89.18 & 87.17 & 88.16 & 78.83 \\
BIT~\citep{chen2021remote} & 2022 & 89.24 & 89.37 & 89.31 & 80.68 \\
RS-Mamba~\citep{zhao2024rsmamba} & 2024 & 91.36 & 88.23 & 89.77 & 81.44 \\
ChangeDA~\citep{meng2025changeda} & 2025 & \textbf{93.67} & \underline{90.92} & \underline{92.27} & \underline{85.65} \\
SPMNet~\citep{wang2025spmnet} & 2025 & 92.12 & 90.58 & 91.34 & 84.07 \\
\hline
\textbf{PhyUnfold-Net (Ours)} & -- & \underline{93.41} & \textbf{92.70} & \textbf{93.05} & \textbf{87.01} \\
\hline
\end{tabular}
}
\end{table}

\begin{table}[!t]
\centering
\caption{Results on \textbf{LEVIR-CD+}. Best in \textbf{bold}, second-best \underline{underlined}.}
\label{tab:levirplus}
\setlength{\tabcolsep}{5pt}
\resizebox{\columnwidth}{!}{%
\begin{tabular}{l|c|cccc}
\hline
Method & Year & Pre.(\%) & Rec.(\%) & F1(\%) & IoU(\%) \\
\hline
FC-Siam-Di~\citep{daudt2018fully} & 2018 & 63.79 & 55.12 & 59.14 & 41.99 \\
DTCDSCN~\citep{liu2020building} & 2021 & 66.38 & 60.51 & 63.31 & 46.30 \\
STANet~\citep{chen2020stanet} & 2020 & 68.72 & 57.43 & 62.58 & 45.53 \\
DSIFN~\citep{zhang2020ifnet} & 2020 & 71.05 & 62.87 & 66.71 & 50.06 \\
SNUNet~\citep{fang2021snunet} & 2022 & 73.91 & 66.23 & 69.86 & 53.69 \\
BIT~\citep{chen2021remote} & 2022 & 73.42 & 68.89 & 71.08 & 55.17 \\
RS-Mamba~\citep{zhao2024rsmamba} & 2024 & 79.67 & 82.19 & 80.91 & 67.95 \\
ChangeDA~\citep{meng2025changeda} & 2025 & 81.47 & 79.85 & 80.65 & 67.58 \\
SPMNet~\citep{wang2025spmnet} & 2025 & \underline{83.59} & \underline{82.60} & \underline{83.09} & \underline{71.01} \\
\hline
\textbf{PhyUnfold-Net (Ours)} & -- & \textbf{87.83} & \textbf{83.89} & \textbf{85.81} & \textbf{75.15} \\
\hline
\end{tabular}
}
\end{table}

\begin{table}[!t]
\centering
\caption{Results on \textbf{WHU-CD}. Best in \textbf{bold}, second-best \underline{underlined}.}
\label{tab:whu}
\setlength{\tabcolsep}{5pt}
\resizebox{\columnwidth}{!}{%
\begin{tabular}{l|c|cccc}
\hline
Method & Year & Pre.(\%) & Rec.(\%) & F1(\%) & IoU(\%) \\
\hline
FC-Siam-Di~\citep{daudt2018fully} & 2018 & 47.33 & 81.53 & 59.88 & 42.73 \\
DTCDSCN~\citep{liu2020building} & 2021 & 63.92 & 82.14 & 71.91 & 56.15 \\
STANet~\citep{chen2020stanet} & 2020 & 79.37 & 78.52 & 78.94 & 65.22 \\
DSIFN~\citep{zhang2020ifnet} & 2020 & 83.40 & 80.13 & 81.73 & 69.10 \\
SNUNet~\citep{fang2021snunet} & 2022 & 85.60 & 87.90 & 86.73 & 76.53 \\
BIT~\citep{chen2021remote} & 2022 & 86.64 & 85.38 & 86.00 & 75.44 \\
RS-Mamba~\citep{zhao2024rsmamba} & 2024 & 95.50 & 90.24 & 92.79 & 86.55 \\
ChangeDA~\citep{meng2025changeda} & 2025 & \underline{96.45} & \underline{91.91} & \underline{94.12} & \underline{88.90} \\
SPMNet~\citep{wang2025spmnet} & 2025 & 94.67 & 89.10 & 91.80 & 84.84 \\
\hline
\textbf{PhyUnfold-Net (Ours)} & -- & \textbf{97.12} & \textbf{94.35} & \textbf{95.71} & \textbf{91.78} \\
\hline
\end{tabular}
}
\end{table}

\begin{table}[!t]
\centering
\caption{Results on \textbf{S2Looking}. Best in \textbf{bold}, second-best \underline{underlined}.}
\label{tab:s2looking}
\setlength{\tabcolsep}{5pt}
\resizebox{\columnwidth}{!}{%
\begin{tabular}{l|c|cccc}
\hline
Method & Year & Pre.(\%) & Rec.(\%) & F1(\%) & IoU(\%) \\
\hline
FC-Siam-Di~\citep{daudt2018fully} & 2018 & 38.25 & 27.64 & 32.09 & 19.11 \\
DTCDSCN~\citep{liu2020building} & 2021 & 42.16 & 31.87 & 36.30 & 22.17 \\
STANet~\citep{chen2020stanet} & 2020 & 43.58 & 33.25 & 37.72 & 23.24 \\
DSIFN~\citep{zhang2020ifnet} & 2020 & 48.33 & 37.42 & 42.17 & 26.72 \\
SNUNet~\citep{fang2021snunet} & 2022 & 50.92 & 40.18 & 44.91 & 28.92 \\
BIT~\citep{chen2021remote} & 2022 & 52.74 & 42.53 & 47.06 & 30.79 \\
RS-Mamba~\citep{zhao2024rsmamba} & 2024 & 67.18 & 58.52 & 62.56 & 45.52 \\
ChangeDA~\citep{meng2025changeda} & 2025 & \textbf{72.26} & \underline{61.45} & \underline{66.42} & \underline{49.72} \\
SPMNet~\citep{wang2025spmnet} & 2025 & 68.92 & 59.37 & 63.78 & 46.85 \\
\hline
\textbf{PhyUnfold-Net (Ours)} & -- & \underline{71.85} & \textbf{63.28} & \textbf{67.29} & \textbf{50.71} \\
\hline
\end{tabular}
}
\end{table}

\noindent\textbf{Analysis.}
Tables~\ref{tab:levir}--\ref{tab:s2looking} summarize quantitative comparisons.
RS-Mamba results are reported on all four benchmarks.
On LEVIR-CD, PhyUnfold-Net achieves the best Rec./F1/IoU, while ChangeDA provides the highest precision.
On LEVIR-CD+, PhyUnfold-Net achieves the best overall scores across all metrics, outperforming both ChangeDA and SPMNet.
On WHU-CD, PhyUnfold-Net achieves the best performance across all four metrics, surpassing ChangeDA by 0.67, 2.44, 1.59, and 2.88 percentage points in Pre., Rec., F1, and IoU, respectively.
On S2Looking, ChangeDA achieves the highest precision (72.26\%).
PhyUnfold-Net achieves the best Rec./F1/IoU (63.28/67.29/50.71), while ChangeDA is second-best on these three metrics.
Overall, PhyUnfold-Net achieves the best F1 and IoU on all four benchmarks among the compared methods, demonstrating robust performance across diverse acquisition conditions.

\subsection{Ablation Study}
\label{subsec:ablation}

We conduct an ablation study on S2Looking. All variants use the same encoder, decoder, and training configuration for a controlled comparison. Table~\ref{tab:ablation} summarizes the contribution of each component.

\begin{table*}[!t]
\centering
\caption{Ablation on \textbf{S2Looking}. $\Delta$ indicates the gain over baseline (\#1).}
\label{tab:ablation}
\setlength{\tabcolsep}{4pt}
\begin{tabular}{c|ccc|cc|cc|cc|cc}
\hline
\# & ICDM & WSSM & S-SEC
& Pre.(\%) & $\Delta$Pre
& Rec.(\%) & $\Delta$Rec
& F1(\%) & $\Delta$F1
& IoU(\%) & $\Delta$IoU \\
\hline
1 & \xmark & \xmark & \xmark & 53.17 & --    & 38.62 & --     & 44.74 & --     & 28.82 & --     \\
2 & \cmark & \xmark & \xmark & 57.63 & +4.46 & 43.78 & +5.16  & 49.76 & +5.02  & 33.12 & +4.30  \\
3 & \xmark & \cmark & \xmark & 59.41 & +6.24 & 45.32 & +6.70  & 51.42 & +6.68  & 34.61 & +5.79  \\
4 & \xmark & \xmark & \cmark & 55.86 & +2.69 & 42.17 & +3.55  & 48.06 & +3.32  & 31.63 & +2.81  \\
5 & \cmark & \cmark & \xmark & 65.28 & +12.11 & 53.74 & +15.12 & 58.95 & +14.21 & 41.79 & +12.97 \\
6 & \cmark & \xmark & \cmark & 62.41 & +9.24 & 50.63 & +12.01 & 55.91 & +11.17 & 38.80 & +9.98  \\
7 & \xmark & \cmark & \cmark & 61.53 & +8.36 & 48.97 & +10.35 & 54.54 & +9.80  & 37.49 & +8.67  \\
8 & \cmark & \cmark & \cmark
& \textbf{71.85} & \textbf{+18.68}
& \textbf{63.28} & \textbf{+24.66}
& \textbf{67.29} & \textbf{+22.55}
& \textbf{50.71} & \textbf{+21.89} \\
\hline
\end{tabular}
\end{table*}

\noindent\textbf{Impact of ICDM.}
ICDM provides the core decomposition capability.
Adding ICDM alone improves the baseline by introducing explicit change--nuisance decomposition.
When combined with WSSM, ICDM raises F1 by 14.21 percentage points (\#5), exceeding the sum of the two standalone F1 gains (5.02 and 6.68 points). This result indicates complementarity in joint change--nuisance disentanglement.

\begin{figure}[!t]
\centering
\includegraphics[width=1.0\linewidth]{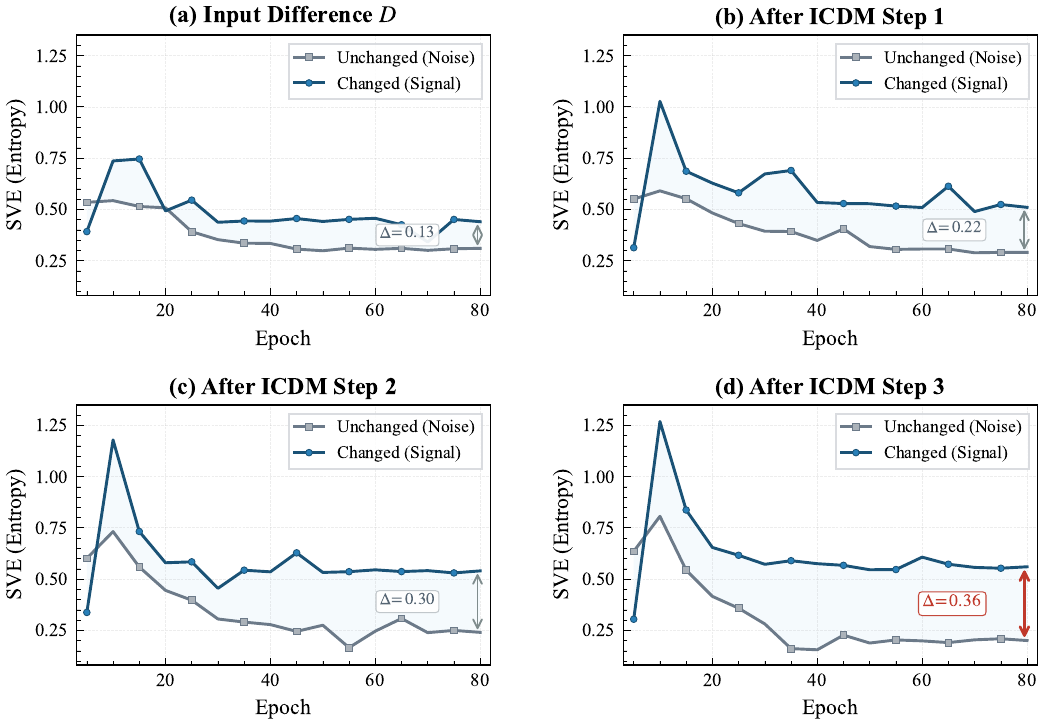}
\caption{Training-time SVE trajectories for changed and unchanged patches on raw $D$ and at ICDM steps $k{=}1,2,3$. The changed-versus-unchanged SVE gap increases from $0.13$ to $0.36$.}
\label{fig:sve_prior}
\end{figure}

\noindent\textbf{Effect of WSSM.}
WSSM is the strongest single module, improving F1 by 6.68 percentage points (\#3).
It also pairs well with ICDM because cleaner inputs make iterative separation more effective under illumination variation.
This synergy suggests that spectral suppression and iterative decomposition address complementary sources of nuisance.

\noindent\textbf{Role of S-SEC.}
S-SEC acts as a stabilization term.
Its standalone F1 gain is 3.32 percentage points (\#4).
Adding S-SEC to the ICDM+WSSM configuration raises F1 by a further 8.34 percentage points (from \#5 to \#8).
The two-phase design reflects the intuition that early steps need freedom to separate components before later steps enforce structural compactness and long-term stability.

\noindent\textbf{ICDM step-wise verification.}
To further validate ICDM, we monitor SVE behavior across unrolling steps during training on LEVIR-CD.
Feature maps at each stage are partitioned into $8{\times}8$ patches,
and SVE is computed per patch grouped by ground-truth labels
(changed and unchanged).

\begin{figure}[!t]
\centering
\includegraphics[width=1.0\linewidth]{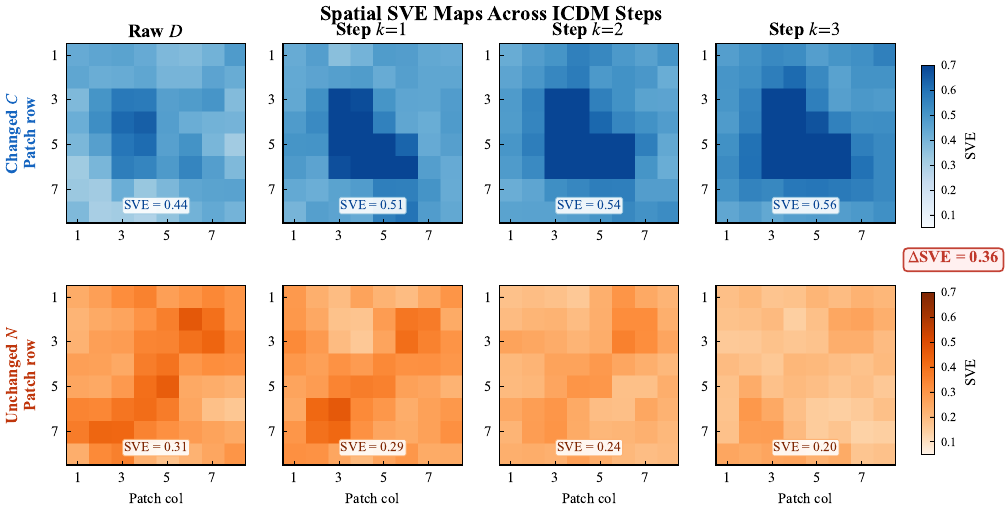}
\caption{Patch-grid SVE maps across ICDM steps. Changed regions become progressively more compact and exhibit higher SVE, whereas unchanged regions become more diffuse and exhibit lower SVE.}
\label{fig:sve_3d}
\end{figure}

Fig.~\ref{fig:sve_prior} reports step-wise SVE trajectories
and Fig.~\ref{fig:sve_3d} reports the corresponding patch maps
across ICDM steps.
We check whether temporal margin growth is stable across steps
and verify whether the same tendency appears in spatial organization.

Agreement between these views reduces the risk of interpreting isolated artifacts as genuine decoupling.
Specifically, progressive decoupling should produce both a widening SVE margin between changed and unchanged patches and increasingly compact changed-region support across ICDM iterations.

\begin{figure}[!t]
\centering
\includegraphics[width=1.0\linewidth]{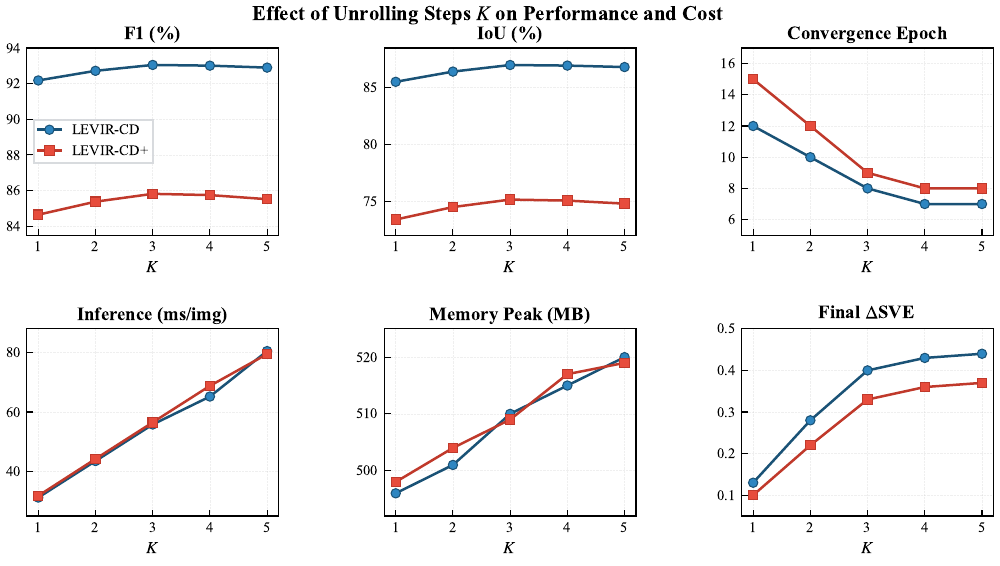}
\caption{Effect of unrolling steps $K$ on LEVIR-CD and LEVIR-CD+.}
\label{fig:k_ablation}
\end{figure}

\begin{figure}[!t]
\centering
\includegraphics[width=1.0\linewidth]{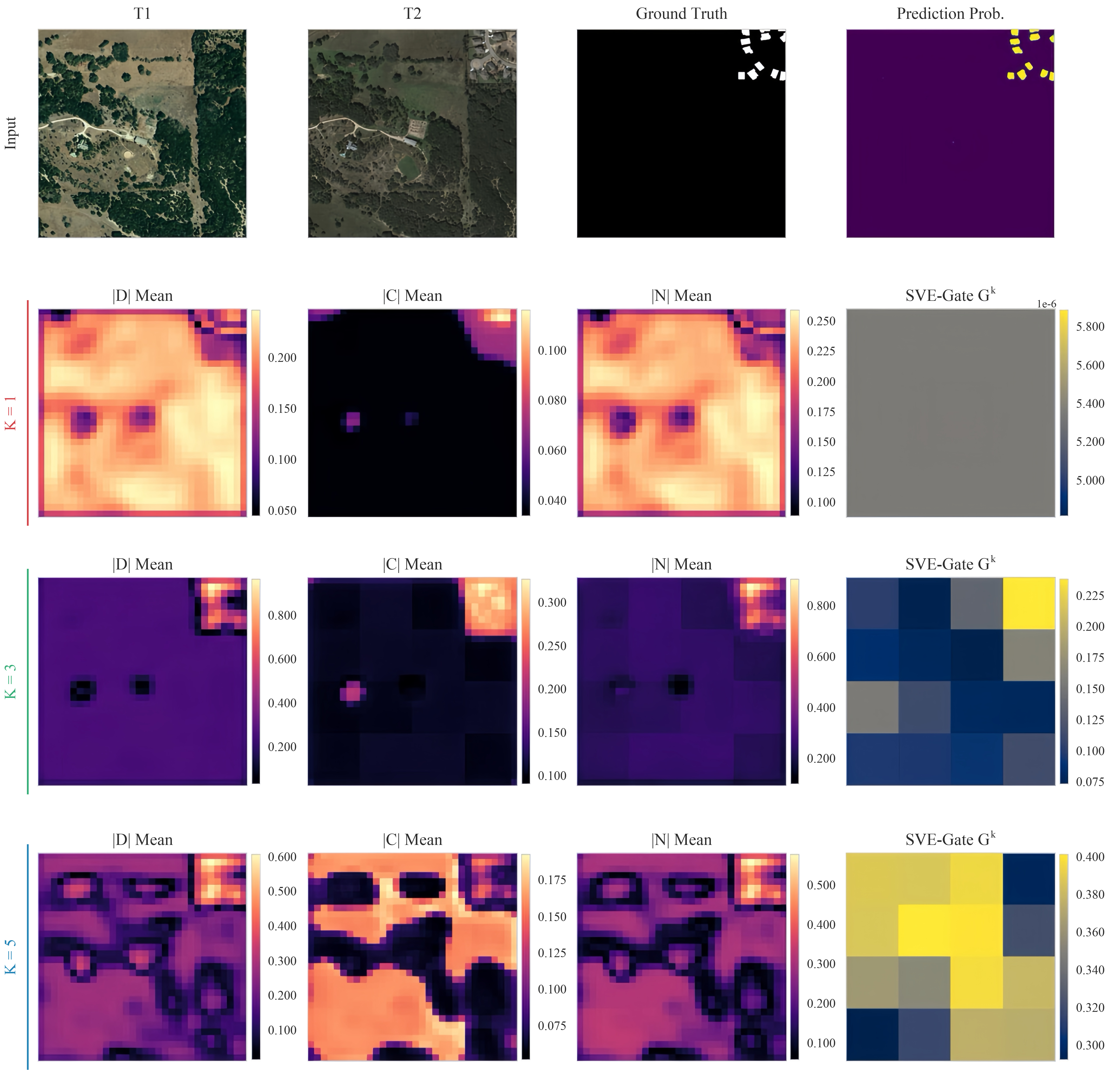}
\caption{Qualitative decomposition at $K{=}1,3,5$ on LEVIR-CD. The result is under-decomposed at $K{=}1$, becomes cleaner at $K{=}3$, and shows only limited additional structural improvement at $K{=}5$.}
\label{fig:k_decomp}
\end{figure}

\noindent\textbf{Effect of unrolling steps $K$.}
On LEVIR-CD and LEVIR-CD+, Fig.~\ref{fig:k_ablation} indicates $K{=}3$ as the best trade-off: accuracy is near saturation, convergence is faster, and incremental $\Delta$SVE gain beyond $K{=}3$ is limited while cost keeps increasing.
After $K{=}3$, extra steps mainly increase runtime and memory while bringing marginal gains.
The gain from $K{=}1$ to $K{=}3$ is larger than from $K{=}3$ to $K{=}5$ for similar added runtime and memory, indicating diminishing returns as mismatch is reduced.
This motivates the qualitative cross-check in Fig.~\ref{fig:k_decomp}, where we examine whether further unrolling beyond $K{=}3$ provides substantial structural gains.

Figure~\ref{fig:k_decomp} visually confirms this trend.
At $K{=}1$, change energy is not fully separated and some structures remain mixed in nuisance responses, whereas at $K{=}3$ the change component better aligns with true changed regions and gate responses become more selective.
At $K{=}5$, the main changed structures are similar to $K{=}3$, but extra textures appear, indicating limited structural benefit with slight over-refinement.
Together with Fig.~\ref{fig:k_ablation}, this supports choosing $K{=}3$ as a practical accuracy--efficiency trade-off.

To examine the empirical contraction trend, we report normalized mismatch scores across ICDM iterations on the LEVIR-CD and S2Looking test sets after training.
For ICDM step $k$, we define the normalized mismatch score as
\[
r^k = \|R_{\mathrm{res}}^k\|_F / \|D\|_F,\qquad R_{\mathrm{res}}^k = D-(C^k+N^k).
\]
The initialization $C^0=\mathbf{0},N^0=D$ gives the trivial residual $r^0=0$. We therefore treat the transition from $k=0$ to $k=1$ as a transient and assess contraction only over $k=1,2,3$.

\begin{table}[!t]
\centering
\caption{Normalized mismatch score $r^k=\|R_{\mathrm{res}}^k\|_F / \|D\|_F$ at ICDM steps $k\in\{1,2,3\}$. ``Ratio'' reports $r^k/r^{k-1}$ for $k\ge 2$.}
\label{tab:residual}
\setlength{\tabcolsep}{6pt}
\begin{tabular}{l|cc|cc}
\toprule
 & \multicolumn{2}{c|}{LEVIR-CD} & \multicolumn{2}{c}{S2Looking} \\
Stage & Score $r^k$ & Ratio & Score $r^k$ & Ratio \\
\midrule
ICDM $k=1$       & 0.412 & --    & 0.487 & --    \\
ICDM $k=2$       & 0.173 & 0.420 & 0.226 & 0.464 \\
ICDM $k=3$       & 0.068 & 0.393 & 0.104 & 0.460 \\
\bottomrule
\end{tabular}
\end{table}

Table~\ref{tab:residual} shows a clear decrease of $r^k$ across ICDM stages.
The similar ratios across successive steps provide empirical evidence of approximately geometric residual decay, but they do not by themselves verify the infinite-step summability assumption.
From the first to the third ICDM step, the mismatch score decreases by $83.5\%$ on LEVIR-CD and $78.6\%$ on S2Looking, supporting why $K{=}3$ is sufficient in practice.

\subsection{Computational Complexity Analysis}
\begin{table}[!t]
\centering
\caption{Results on \textbf{LEVIR-CD+} with computational complexity. Best results are in \textbf{bold}, and second-best results are \underline{underlined}.}
\label{tab:levirplus_flops}
\setlength{\tabcolsep}{6pt}
\resizebox{\columnwidth}{!}{%
\begin{tabular}{@{}l|c|ccc|c@{}}
\toprule[1.15pt]
Method & Year & Pre.(\%) & F1(\%) & IoU(\%) & FLOPs (G) \\
\midrule
STANet~\citep{chen2020stanet}      & 2020 & 68.72 & 62.58 & 45.53 & 3706.81 \\
DSIFN~\citep{zhang2020ifnet}       & 2020 & 71.05 & 66.71 & 50.06 & 2632.16 \\
SNUNet~\citep{fang2021snunet}      & 2022 & 73.91 & 69.86 & 53.69 & 1754.66 \\
RS-Mamba~\citep{zhao2024rsmamba}   & 2024 & 79.67 & 80.91 & 67.95 & \underline{827.83} \\
ChangeDA~\citep{meng2025changeda}  & 2025 & 81.47 & 80.65 & 67.58 & 1848.49 \\
SPMNet~\citep{wang2025spmnet}      & 2025 & \underline{83.59} & \underline{83.09} & \underline{71.01} & 2388.91 \\
\midrule
\textbf{PhyUnfold-Net}     & -- & \textbf{87.83} & \textbf{85.81} & \textbf{75.15} & \textbf{438.06} \\
\bottomrule[1.15pt]
\end{tabular}
}
\end{table}
As shown in Table~\ref{tab:levirplus_flops}, our method achieves a higher IoU than SPMNet while requiring substantially fewer FLOPs.
This result demonstrates that the complete PhyUnfold-Net architecture provides a favorable accuracy--complexity trade-off; it does not isolate the computational cost of ICDM alone.

\subsection{Hyperparameter Sensitivity}
\noindent\textbf{Separation Margin:} Table~\ref{tab:sens_m} shows that $m{=}0.1$ provides insufficient separation pressure, whereas $m{=}0.5$ over-constrains the representation space and reduces recall.
Performance peaks at $m{=}0.3$ and remains stable over $[0.2,0.4]$.

\begin{table}[!t]
\centering
\caption{Sensitivity to separation margin $m$ on S2Looking.}
\label{tab:sens_m}
\setlength{\tabcolsep}{8pt}
\begin{tabular}{c|cccc}
\toprule
$m$ & Pre.(\%) & Rec.(\%) & F1(\%) & IoU(\%) \\
\midrule
0.1 & 67.83 & 61.42 & 64.47 & 47.56 \\
0.2 & 70.29 & 62.85 & 66.36 & 49.66 \\
\textbf{0.3} & \textbf{71.85} & \textbf{63.28} & \textbf{67.29} & \textbf{50.71} \\
0.4 & 71.56 & 61.93 & 66.40 & 49.70 \\
0.5 & 69.14 & 60.78 & 64.69 & 47.81 \\
\bottomrule
\end{tabular}
\end{table}

\noindent\textbf{Nuisance Energy Bounds:} As shown in Table~\ref{tab:sens_tau}, the tight pair $(\tau_\ell{=}0.01,\tau_u{=}0.20)$ yields lower segmentation accuracy, consistent with an overly restrictive nuisance representation.
The loose pair $(\tau_\ell{=}0.10,\tau_u{=}0.60)$ also reduces performance, indicating insufficient control over nuisance energy. The default pair $(0.05,0.40)$ provides the best trade-off.

\begin{table}[!t]
\centering
\caption{Sensitivity to nuisance bounds $(\tau_\ell, \tau_u)$ on S2Looking.}
\label{tab:sens_tau}
\setlength{\tabcolsep}{6pt}
\begin{tabular}{cc|cccc}
\toprule
$\tau_\ell$ & $\tau_u$ & Pre.(\%) & Rec.(\%) & F1(\%) & IoU(\%) \\
\midrule
0.01 & 0.20 & 68.97 & 60.35 & 64.37 & 47.46 \\
0.03 & 0.30 & 70.18 & 62.54 & 66.14 & 49.41 \\
\textbf{0.05} & \textbf{0.40} & \textbf{71.85} & \textbf{63.28} & \textbf{67.29} & \textbf{50.71} \\
0.08 & 0.50 & 71.39 & 62.17 & 66.46 & 49.77 \\
0.10 & 0.60 & 69.82 & 61.89 & 65.62 & 48.83 \\
\bottomrule
\end{tabular}
\end{table}

\noindent\textbf{Loss Weights:} We report the metrics with different loss weights in Table~\ref{tab:sens_lambda}. When both weights are small ($\lambda_{\mathrm{e}}=0.1, \lambda_{\mathrm{c}}=0.5$), S-SEC provides weak regularization and decomposition quality degrades.
Overly strong regularization ($\lambda_{\mathrm{e}}=1.0, \lambda_{\mathrm{c}}=2.0$) competes with $\mathcal{L}_{\mathrm{seg}}$ and harms segmentation accuracy.
A moderate setting, such as $\lambda_{\mathrm{e}}=0.5$ and $\lambda_{\mathrm{c}}=1.0$, yields the best performance.
Increasing $\lambda_{\mathrm{c}}$ from 1.0 to 2.0 at fixed $\lambda_{\mathrm{e}}=0.5$ causes only a modest F1 decrease, indicating limited sensitivity around the selected setting.

\begin{table}[!t]
\centering
\caption{Sensitivity to loss weights $(\lambda_{\mathrm{e}}, \lambda_{\mathrm{c}})$ on S2Looking.}
\label{tab:sens_lambda}
\setlength{\tabcolsep}{6pt}
\begin{tabular}{cc|cccc}
\toprule
$\lambda_{\mathrm{e}}$ & $\lambda_{\mathrm{c}}$ & Pre.(\%) & Rec.(\%) & F1(\%) & IoU(\%) \\
\midrule
0.1 & 0.5 & 68.93 & 61.07 & 64.76 & 47.89 \\
0.3 & 0.5 & 70.46 & 62.73 & 66.37 & 49.67 \\
\textbf{0.5} & \textbf{1.0} & \textbf{71.85} & \textbf{63.28} & \textbf{67.29} & \textbf{50.71} \\
0.5 & 2.0 & 71.68 & 62.41 & 66.72 & 50.07 \\
1.0 & 1.0 & 70.51 & 62.19 & 66.09 & 49.35 \\
1.0 & 2.0 & 69.37 & 60.54 & 64.65 & 47.77 \\
\bottomrule
\end{tabular}
\end{table}

\FloatBarrier

\section{Conclusion}
\label{sec:conclusion}

In this paper, we present PhyUnfold-Net, a physics-guided deep unfolding framework for bi-temporal change detection via explicit difference-space decomposition.
The model integrates ICDM for change--nuisance separation, staged S-SEC regularization for stable decomposition, and WSSM for suppressing acquisition-induced spectral mismatch.
Experiments on four benchmarks demonstrate strong and consistent performance, with PhyUnfold-Net achieving the best F1 and IoU on all four datasets among the methods compared.
Future work will study spatially adaptive entropy estimation and extend to multi-temporal change detection.

\clearpage
\bibliographystyle{unsrtnat}
\bibliography{mybib}

@article{beck2009fista,
  title   = {A Fast Iterative Shrinkage-Thresholding Algorithm for Linear Inverse Problems},
  author  = {Beck, Amir and Teboulle, Marc},
  journal = {SIAM Journal on Imaging Sciences},
  volume  = {2},
  number  = {1},
  pages   = {183--202},
  year    = {2009},
  doi     = {10.1137/080716542}
}

@inproceedings{gregor2010learning,
  title     = {Learning Fast Approximations of Sparse Coding},
  author    = {Gregor, Karol and LeCun, Yann},
  booktitle = {Proceedings of the 27th International Conference on Machine Learning (ICML)},
  pages     = {399--406},
  year      = {2010}
}

@inproceedings{cho2014learning,
  title     = {Learning Phrase Representations using {RNN} Encoder-Decoder for Statistical Machine Translation},
  author    = {Cho, Kyunghyun and van Merrienboer, Bart and Gulcehre, Caglar and Bahdanau, Dzmitry and Bougares, Fethi and Schwenk, Holger and Bengio, Yoshua},
  booktitle = {Proceedings of the 2014 Conference on Empirical Methods in Natural Language Processing (EMNLP)},
  pages     = {1724--1734},
  year      = {2014},
  doi       = {10.3115/v1/D14-1179}
}

@inproceedings{ronneberger2015unet,
  title     = {{U-Net}: Convolutional Networks for Biomedical Image Segmentation},
  author    = {Ronneberger, Olaf and Fischer, Philipp and Brox, Thomas},
  booktitle = {Medical Image Computing and Computer-Assisted Intervention (MICCAI)},
  pages     = {234--241},
  year      = {2015},
  doi       = {10.1007/978-3-319-24574-4_28}
}

@inproceedings{shi2015convolutional,
  title     = {Convolutional {LSTM} Network: A Machine Learning Approach for Precipitation Nowcasting},
  author    = {Shi, Xingjian and Chen, Zhourong and Wang, Hao and Yeung, Dit-Yan and Wong, Wai-Kin and Woo, Wang-Chun},
  booktitle = {Advances in Neural Information Processing Systems (NeurIPS)},
  volume    = {28},
  pages     = {802--810},
  year      = {2015}
}

@inproceedings{he2016deep,
  title     = {Deep Residual Learning for Image Recognition},
  author    = {He, Kaiming and Zhang, Xiangyu and Ren, Shaoqing and Sun, Jian},
  booktitle = {Proceedings of the IEEE Conference on Computer Vision and Pattern Recognition (CVPR)},
  pages     = {770--778},
  year      = {2016},
  doi       = {10.1109/CVPR.2016.90}
}

@inproceedings{milletari2016vnet,
  title     = {{V-Net}: Fully Convolutional Neural Networks for Volumetric Medical Image Segmentation},
  author    = {Milletari, Fausto and Navab, Nassir and Ahmadi, Seyed-Ahmad},
  booktitle = {Proceedings of the International Conference on 3D Vision (3DV)},
  pages     = {565--571},
  year      = {2016},
  doi       = {10.1109/3DV.2016.79}
}

@inproceedings{deng2009imagenet,
  title     = {{ImageNet}: A Large-Scale Hierarchical Image Database},
  author    = {Deng, Jia and Dong, Wei and Socher, Richard and Li, Li-Jia and Li, Kai and Fei-Fei, Li},
  booktitle = {Proceedings of the IEEE Conference on Computer Vision and Pattern Recognition (CVPR)},
  pages     = {248--255},
  year      = {2009},
  doi       = {10.1109/CVPR.2009.5206848}
}

@inproceedings{lin2017feature,
  title     = {Feature Pyramid Networks for Object Detection},
  author    = {Lin, Tsung-Yi and Dollar, Piotr and Girshick, Ross and He, Kaiming and Hariharan, Bharath and Belongie, Serge},
  booktitle = {Proceedings of the IEEE Conference on Computer Vision and Pattern Recognition (CVPR)},
  pages     = {936--944},
  year      = {2017},
  doi       = {10.1109/CVPR.2017.106}
}

@inproceedings{vaswani2017attention,
  title     = {Attention Is All You Need},
  author    = {Vaswani, Ashish and Shazeer, Noam and Parmar, Niki and Uszkoreit, Jakob and Jones, Llion and Gomez, Aidan N. and Kaiser, Lukasz and Polosukhin, Illia},
  booktitle = {Advances in Neural Information Processing Systems (NeurIPS)},
  volume    = {30},
  pages     = {5998--6008},
  year      = {2017}
}

@inproceedings{zhang2018ista,
  title     = {{ISTA-Net}: Interpretable Optimization-Inspired Deep Network for Image Compressive Sensing},
  author    = {Zhang, Jian and Ghanem, Bernard},
  booktitle = {Proceedings of the IEEE Conference on Computer Vision and Pattern Recognition (CVPR)},
  pages     = {1828--1837},
  year      = {2018},
  doi       = {10.1109/CVPR.2018.00196}
}

@inproceedings{daudt2018fully,
  title     = {Fully Convolutional Siamese Networks for Change Detection},
  author    = {Daudt, Rodrigo Caye and Le Saux, Bertr and Boulch, Alexandre},
  booktitle = {Proceedings of the IEEE International Conference on Image Processing (ICIP)},
  pages     = {4063--4067},
  year      = {2018},
  doi       = {10.1109/ICIP.2018.8451652}
}

@article{ji2018fully,
  title   = {Fully Convolutional Networks for Multisource Building Extraction From an Open Aerial and Satellite Imagery Data Set},
  author  = {Ji, Shunping and Wei, Shiqing and Lu, Meng},
  journal = {IEEE Transactions on Geoscience and Remote Sensing},
  volume  = {57},
  number  = {1},
  pages   = {574--586},
  year    = {2019},
  doi     = {10.1109/TGRS.2018.2858817}
}

@article{loshchilov2019decoupled,
  title   = {Decoupled Weight Decay Regularization},
  author  = {Loshchilov, Ilya and Hutter, Frank},
  journal = {International Conference on Learning Representations (ICLR)},
  year    = {2019}
}

@inproceedings{kingma2015adam,
  title     = {Adam: A Method for Stochastic Optimization},
  author    = {Kingma, Diederik P. and Ba, Jimmy},
  booktitle = {International Conference on Learning Representations (ICLR)},
  year      = {2015}
}

@article{liu2020building,
  title   = {Building Change Detection for Remote Sensing Images Using a Dual-Task Constrained Deep Siamese Convolutional Network Model},
  author  = {Liu, Yi and Pang, Chao and Zhan, Zongqian and Zhang, Xiaomeng and Yang, Xue},
  journal = {IEEE Geoscience and Remote Sensing Letters},
  volume  = {18},
  number  = {5},
  pages   = {811--815},
  year    = {2021},
  doi     = {10.1109/LGRS.2020.2988032}
}

@article{zhang2020ifnet,
  title   = {A Deeply Supervised Image Fusion Network for Change Detection in High Resolution Bi-Temporal Remote Sensing Images},
  author  = {Zhang, Chenxiao and Yue, Peng and Tapete, Deodato and Jiang, Liangcun and Shangguan, Boyi and Huang, Li and Liu, Guangchao},
  journal = {ISPRS Journal of Photogrammetry and Remote Sensing},
  volume  = {166},
  pages   = {183--200},
  year    = {2020},
  doi     = {10.1016/j.isprsjprs.2020.06.003}
}

@article{chen2020stanet,
  title   = {A Spatial-Temporal Attention-Based Method and a New Dataset for Remote Sensing Image Change Detection},
  author  = {Chen, Hao and Shi, Zhenwei},
  journal = {Remote Sensing},
  volume  = {12},
  number  = {10},
  pages   = {1662},
  year    = {2020},
  doi     = {10.3390/rs12101662}
}

@article{fang2021snunet,
  title   = {{SNUNet-CD}: A Densely Connected Siamese Network for Change Detection of {VHR} Images},
  author  = {Fang, Sheng and Li, Kaiyu and Shao, Jinyuan and Li, Zhe},
  journal = {IEEE Geoscience and Remote Sensing Letters},
  volume  = {19},
  pages   = {1--5},
  year    = {2022},
  doi     = {10.1109/LGRS.2021.3056416}
}

@article{chen2021remote,
  title   = {Remote Sensing Image Change Detection With Transformers},
  author  = {Chen, Hao and Qi, Zipeng and Shi, Zhenwei},
  journal = {IEEE Transactions on Geoscience and Remote Sensing},
  volume  = {60},
  pages   = {1--14},
  year    = {2022},
  doi     = {10.1109/TGRS.2021.3095166}
}

@article{zhao2024rsmamba,
  title   = {{RS-Mamba} for Large Remote Sensing Image Dense Prediction},
  author  = {Zhao, Sijie and Chen, Hao and Zhang, Xueliang and Xiao, Pengfeng and Bai, Lei and Ouyang, Wanli},
  journal = {IEEE Transactions on Geoscience and Remote Sensing},
  volume  = {62},
  pages   = {1--14},
  year    = {2024},
  doi     = {10.1109/TGRS.2024.3425540}
}

@article{shi2022deeply,
  title   = {A Deeply Supervised Attention Metric-Based Network and an Open Aerial Image Dataset for Remote Sensing Change Detection},
  author  = {Shi, Qian and Liu, Mengxi and Li, Shengchen and Liu, Xiaoping and Wang, Fei and Zhang, Liangpei},
  journal = {IEEE Transactions on Geoscience and Remote Sensing},
  volume  = {60},
  pages   = {1--16},
  year    = {2022},
  doi     = {10.1109/TGRS.2021.3085870}
}

@article{han2023hanet,
  title   = {{HANet}: A Hierarchical Attention Network for Change Detection With Bitemporal Very-High-Resolution Remote Sensing Images},
  author  = {Han, Chengxi and Wu, Chen and Guo, Haonan and Hu, Meiqi and Chen, Hongruixuan},
  journal = {IEEE Journal of Selected Topics in Applied Earth Observations and Remote Sensing},
  volume  = {16},
  pages   = {3867--3878},
  year    = {2023},
  doi     = {10.1109/JSTARS.2023.3264802}
}

@inproceedings{chen2023saras,
  title     = {{SARAS-Net}: Scale and Relation Aware Siamese Network for Change Detection},
  author    = {Chen, Chao-Peng and Wu, Jun-Wei and Chen, Lyu-Guang and Wan, Wei},
  booktitle = {Proceedings of the AAAI Conference on Artificial Intelligence},
  volume    = {37},
  pages     = {14187--14195},
  year      = {2023},
  doi       = {10.1609/aaai.v37i12.26660}
}

@article{codegoni2023tinycd,
  title   = {{TinyCD}: A (Not So) Deep Learning Model for Change Detection},
  author  = {Codegoni, Andrea and Lombardi, Gabriele and Ferrari, Alessandro},
  journal = {Neural Computing and Applications},
  volume  = {35},
  pages   = {8471--8486},
  year    = {2023},
  doi     = {10.1007/s00521-022-08122-3}
}

@article{shen2021s2looking,
  title   = {{S2Looking}: A Satellite Side-Looking Dataset for Building Change Detection},
  author  = {Shen, Li and Lu, Yao and Chen, Hao and Wei, Hao and Xie, Donghai and Yue, Jiabao and Chen, Rui and Lv, Shouye and Jiang, Bitao},
  journal = {Remote Sensing},
  volume  = {13},
  number  = {24},
  pages   = {5094},
  year    = {2021},
  doi     = {10.3390/rs13245094}
}

@inproceedings{dosovitskiy2021image,
  title     = {An Image Is Worth 16x16 Words: Transformers for Image Recognition at Scale},
  author    = {Dosovitskiy, Alexey and Beyer, Lucas and Kolesnikov, Alexander and Weissenborn, Dirk and Zhai, Xiaohua and Unterthiner, Thomas and Dehghani, Mostafa and Minderer, Matthias and Heigold, Georg and Gelly, Sylvain and Uszkoreit, Jakob and Houlsby, Neil},
  booktitle = {International Conference on Learning Representations (ICLR)},
  year      = {2021}
}

@inproceedings{liu2021swin,
  title     = {{Swin Transformer}: Hierarchical Vision Transformer using Shifted Windows},
  author    = {Liu, Ze and Lin, Yutong and Cao, Yue and Hu, Han and Wei, Yixuan and Zhang, Zheng and Lin, Stephen and Guo, Baining},
  booktitle = {Proceedings of the IEEE/CVF International Conference on Computer Vision (ICCV)},
  pages     = {9992--10002},
  year      = {2021},
  doi       = {10.1109/ICCV48922.2021.00986}
}

@inproceedings{xie2021segformer,
  title     = {{SegFormer}: Simple and Efficient Design for Semantic Segmentation With Transformers},
  author    = {Xie, Enze and Wang, Wenhai and Yu, Zhiding and Anandkumar, Anima and Alvarez, Jose M. and Luo, Ping},
  booktitle = {Advances in Neural Information Processing Systems (NeurIPS)},
  volume    = {34},
  pages     = {12077--12090},
  year      = {2021}
}

@article{monga2021algorithm,
  title   = {Algorithm Unrolling: Interpretable, Efficient Deep Learning for Signal and Image Processing},
  author  = {Monga, Vishal and Li, Yuelong and Eldar, Yonina C.},
  journal = {IEEE Signal Processing Magazine},
  volume  = {38},
  number  = {2},
  pages   = {18--44},
  year    = {2021},
  doi     = {10.1109/MSP.2020.3016905}
}

@article{ding2024joint,
  title   = {Joint Spatio-Temporal Modeling for Semantic Change Detection in Remote Sensing Images},
  author  = {Ding, Lei and Guo, Haitao and Liu, Sicong and Mou, Lichao and Zhang, Jing and Bruzzone, Lorenzo},
  journal = {IEEE Transactions on Geoscience and Remote Sensing},
  volume  = {62},
  pages   = {1--14},
  year    = {2024},
  doi     = {10.1109/TGRS.2024.3362795}
}

@article{meng2025changeda,
  title   = {{ChangeDA}: Depth-Augmented Multitask Network for Remote Sensing Change Detection via Differential Analysis},
  author  = {Meng, Jiangtao and Xu, Xinying and Zhang, Zhe and Li, Pengyue and Xie, Gang and Ren, Jinchang and Zheng, Yuxuan},
  journal = {IEEE Transactions on Geoscience and Remote Sensing},
  volume  = {63},
  pages   = {1--19},
  year    = {2025},
  doi     = {10.1109/TGRS.2025.3532468}
}

@article{wang2025spmnet,
  title   = {{SPMNet}: A Siamese Pyramid {Mamba} Network for Very-High-Resolution Remote Sensing Change Detection},
  author  = {Wang, Jiashu and Song, Jinze and Zhang, Hao and Zhang, Zekai and Ji, Yunlong and Zhang, Wenyin and Zhang, Jinglin and Wang, Xing},
  journal = {IEEE Transactions on Geoscience and Remote Sensing},
  volume  = {63},
  pages   = {1--14},
  year    = {2025},
  doi     = {10.1109/TGRS.2025.3565801}
}

\end{document}